\documentclass[journal]{IEEEtran}
\usepackage{booktabs}
\usepackage{amsmath,amssymb,amsfonts}
\usepackage{algorithmicx,algorithm}
\usepackage{algorithm,algpseudocode}
\usepackage{multirow,bbding}
\usepackage{xcolor}
\usepackage{booktabs}
\usepackage[normalem]{ulem}
\usepackage{soul}
\newcommand{\eat}[1]{}
\usepackage{booktabs}
\usepackage{hyperref}
\usepackage{cite}
\usepackage{colortbl}

\ifCLASSINFOpdf
\usepackage[pdftex]{graphicx}
\else
\fi

\begin{document}
	\title{Continual Referring Expression
		Comprehension via Dual Modular Memorization}
		\author{Heng Tao Shen~\IEEEmembership{Fellow, IEEE}, Cheng Chen, Peng Wang, Lianli Gao, Jingkuan Song, Meng Wang~\IEEEmembership{Fellow, IEEE}
		\thanks{Heng Tao Shen and Jingkuan Song are with Center for Future Media, University of Electronic Science and Technology of China and Peng Cheng Laboratory.
		Cheng Chen, Lianli Gao  are with the School of Computer Science, University of Electronic Science and Technology of China, China.
		Peng Wang is with the School of Computing and Information Technology, University of Wollongong. Meng Wang is with School of Computer Science and Information Engineering, Hefei University of Technology, China. \textit{Corresponding author: Jingkuan Song.}}
		}
	\eat{\author{Michael~Shell,~\IEEEmembership{Member,~IEEE,}
			John~Doe,~\IEEEmembership{Fellow,~OSA,}
			and~Jane~Doe,~\IEEEmembership{Life~Fellow,~IEEE}
			\thanks{M. Shell was with the Department
				of Electrical and Computer Engineering, Georgia Institute of Technology, Atlanta,
				GA, 30332 USA e-mail: (see http://www.michaelshell.org/contact.html).}
			\thanks{J. Doe and J. Doe are with Anonymous University.}
			\thanks{Manuscript received April 19, 2005; revised August 26, 2015.}}}
	
	\markboth{Journal of \LaTeX\ Class Files,~Vol.~14, No.~8, August~2015}%
	{Shell \MakeLowercase{\textit{et al.}}: Bare Demo of IEEEtran.cls for IEEE Journals}
	
	\maketitle
	
	\begin{abstract}
		Referring Expression Comprehension (REC) aims to localize an image region of a given object described by a natural-language expression. While promising performance has been demonstrated, existing REC algorithms make a strong assumption that training data feeding into a model are given upfront, which degrades its practicality for real-world scenarios. In this paper, we propose Continual Referring Expression Comprehension (CREC), a new setting for REC, where a model is learning on a stream of incoming tasks. In order to continuously improve the model on sequential tasks without forgetting prior learned knowledge and without repeatedly re-training from a scratch, we propose an effective baseline method named Dual Modular Memorization (DMM), which alleviates the problem of catastrophic forgetting by two memorization modules: Implicit-Memory and Explicit-Memory.
		Specifically, the former module aims to constrain drastic changes to important parameters learned on old tasks when learning a new task; while the latter module maintains a buffer pool to dynamically select and store representative samples of each seen task for future rehearsal. We create three benchmarks for the new CREC setting, by respectively re-splitting three widely-used REC datasets RefCOCO, RefCOCO+ and RefCOCOg into sequential tasks. Extensive experiments on the constructed benchmarks demonstrate that our DMM method significantly outperforms  other  alternatives, based on two  popular REC backbones.
		We make the source code and benchmarks publicly available to foster future progress in this field: \url{https://github.com/zackschen/DMM}.
		
	\end{abstract}
	
	
	\IEEEpeerreviewmaketitle

	\section{Introduction}
	
	\IEEEPARstart{R}{eferring} Expression Comprehension (REC) (or visual grounding) \cite{yu2018mattnet,DBLP:journals/tip/RongYT20,DBLP:journals/tip/LiuWWY20} aims to localize an image region of an object described by a natural-language expression.
	With the increased interest in human-computer communication, REC has been widely applied to various downstream tasks, including image retrieval \cite{DBLP:conf/eccv/LeeCHHH18,wang2019position,DBLP:journals/tip/FuhCE00,DBLP:journals/tip/ZhangZ07}, visual question answering \cite{gao2019structured,zhang2019interpretable,gao2021hierarchical} and language based navigation \cite{tan2019learning,zhu2020vision}. 

	{Over the years, REC models have been improved in several ways.} 
	Early REC works \cite{yu2016modeling,mao2016generation,hu2016natural} use CNN-LSTM frameworks to find the referred region. \cite{chen2017query,luo2017comprehension,rohrbach2016grounding} treats REC as a cross-domain matching problem.
	By introducing modular networks to handle expressions with different types of information, recently proposed \cite{yu2018mattnet} and \cite{liu2019improving} have remarkably advanced the state-of-the-art in REC performance. 
	But all of them are based on an assumption that training data feeding into a model are given upfront.
	This needs expensive training data annotations.
\begin{figure}[t]
		\centering
		\includegraphics[width=0.5\textwidth]{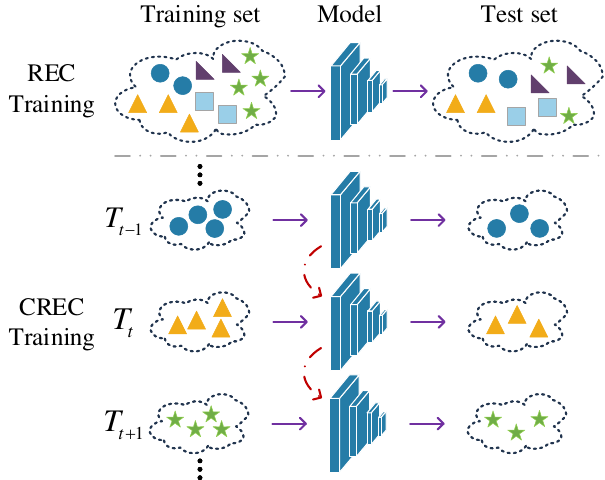}
		\caption{Comparison between REC and CREC.  
			Different from REC that utilizes all samples of the training set to train a model at once, CREC considers a sequential setting where subsets of the training set (i.e., tasks) are revealed one after another. Different colors indicate different groups of tasks, the subscript represents the order of training.} 
		\label{fig:REC_setting}
	\end{figure}
	
	This view of supervised learning stands in contrast with how humans acquire knowledge.
	In real-world scenarios, the setting is more complex and challenging. 
	A model needs to learn from a stream of data instead of all the samples which have been collected completely.
	Note that, a more significant challenge is that during training on the stream, the training data from previous are unavailable.
	{This type of learning is referred to as continual learning (sometimes incremental or lifelong learning).}
	
	\begin{figure*}[t]
		\centering
		\includegraphics[width=\textwidth]{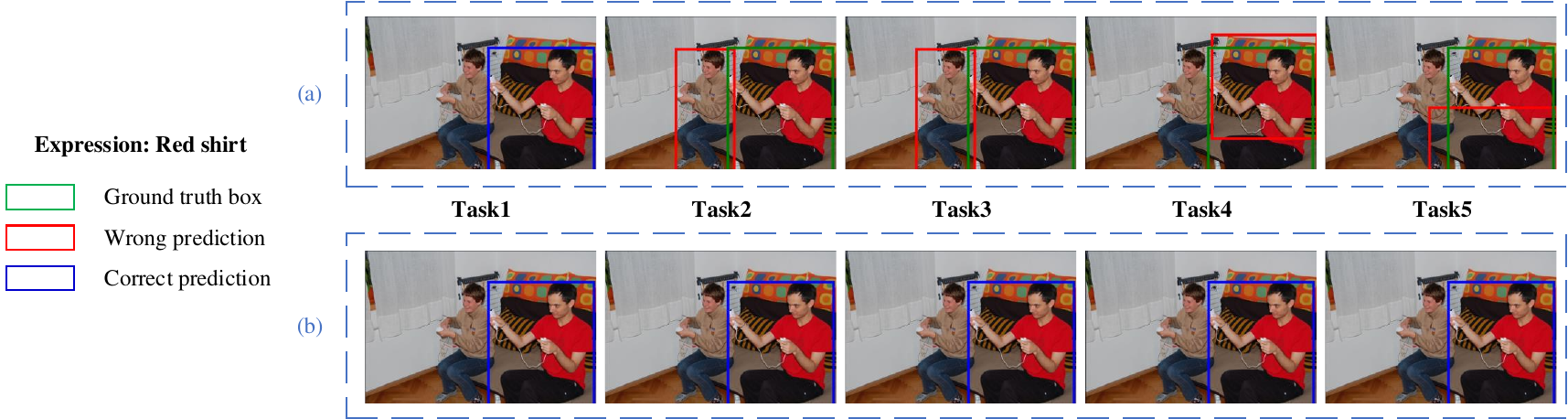}
		\caption{Qualitative grounding results for (a) MAttNet and (b) DMM on the proposed CREC benchmark dataset CRefCOCO (see Section \ref{datasplit}), under the 5-task setting. 
			For each method, we report the test grounding results on the first task once the model is sufficiently trained on a new task.  
			Results clearly show that MAttNet cannot avoid the problem of catastrophic forgetting, but the designed DMM can.}
		\label{fig:catforgetting}
	\end{figure*}
	
	Continual learning \cite{silver2013lifelong,zenke2017continual} is such a practical learning paradigm that divides the stream of data into multiple tasks according to the characteristics of the data and considers a sequential learning setting where tasks are revealed to a model one after another. 
	The number of samples per task may be imbalanced.
	One consequence of learning under such a setting is that as a model learns new tasks, its performance on old ones degrades. This phenomenon is known as ``catastrophic forgetting'' \cite{goodfellow2013empirical,kirkpatrick2017overcoming}, which caused by the dilemma of {stability-plasticity}. 
	In concrete terms, {plasticity} indicates a model's ability to learn new knowledge, while {stability} represents the model's capacity to retain prior knowledge.

	While the promising performance of REC has been demonstrated, it still has a long way to go before it can be practically applied to real application. Firstly, the classical supervised REC learning systems acquire knowledge by providing them with a large number of annotated training samples. This view of supervised learning stands in stark contrast with how humans acquire knowledge. 
	Secondly, in practical applications, we start training once we obtain data. We cannot wait until a large amount of data arrives before training. 
	The REC models must sometimes be updated on-the-fly to recognize new concepts, while the training data are sometimes unavailable for reuse. Thirdly, collecting such a large number of training samples requires a lot of manual effort. And the noises of samples will inevitably occur. To solve these problems, we can collect just a portion of samples to train a model. Then we update this model when we collect another portion of samples. 
	So, in order to improve the practicality of REC,
	in this paper, we propose a novel task of Continual Referring Expression  Comprehension (CREC) to improve the practicality of REC in real-world scenarios. 
	Different from the standard REC task, in which the model is trained only once on a static training set,
	CREC considers a continual learning setting where the training data arrives in a streaming fashion, as depicted in Fig. \ref{fig:REC_setting}. 
	Because REC is the upstream task for visual question answering and visual dialog, the CREC can facilitate the process of applying these methods.

\begin{figure*}[t]
		\centering
		\includegraphics[width=0.97\textwidth]{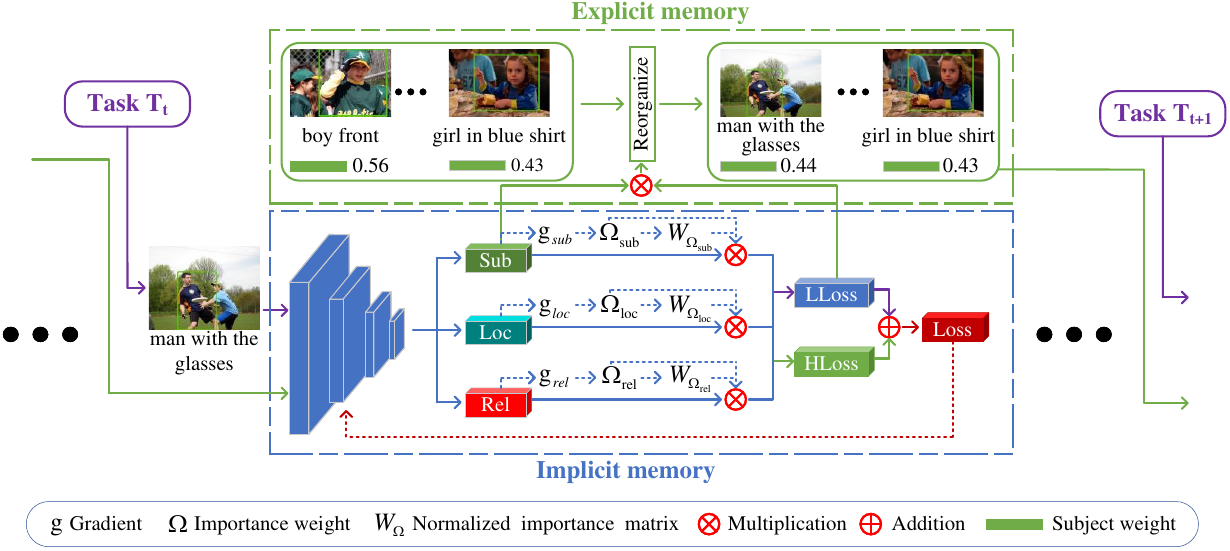}
		\caption{Illustration of the proposed \textbf{Dual Modular Memorization (DMM)} network. 
			DMM seeks to avoid the catastrophic forgetting problem when learning a new task, with the  following appealing components: 
			{\textbf{(1) Implicit-Memory}}, which is built on a modular attention network for decomposing the model into three sub-modules related to \textit{subject} (\textit{Sub}), \textit{location} (\textit{Loc}) and \textit{relationship} (\textit{Rel}). 
			We compute the importance weight of the intra-module parameters $\Omega$ by the gradient $g$ and normalize the inter-module importance matrix $W$ to avoid dramatic changes to parameters learned on the previous tasks.
			{\textbf{(2) Explicit-Memory}}, seeking to maintain a buffer pool to dynamically select and store representative samples of each seen task for future rehearsal. 
			When learning a new task, the model is jointly optimized on samples from the current task and the buffer pool.
		}
		\label{fig:DMM}
	\end{figure*}

	While the new CREC setting affords better practicality, existing REC models, under this setting, suffer from catastrophic forgetting when sequentially trained on a series of tasks (as shown in Fig. \ref{fig:catforgetting} (a)).
	Furthermore, existing continual learning methods are usually designed for image classification tasks, which may neglect the intrinsic characteristics of CREC task. For example, different CREC tasks may rely on different aspects of module, e.g., subject, location and relationship as shown in Fig.~\ref{fig:DMM}. Existing methods usually treat the parameters as isolated elements and ignore the modular information. Therefore, an inferior performance is usually achieved if we directly apply existing continual learning models to the CREC task.
	To address this issue, we develop a novel Dual Modular Memorization (DMM) mechanism for this new CREC task, which consists of two key modules of \textit{Implicit-Memory} and \textit{Explicit-Memory}, as depicted in Fig. \ref{fig:DMM}.
	
	Specifically, the Implicit-Memory module is built upon a standard modular attention network (e.g, MAttNet or CM-Att-Erase). 
	Inspired by Memory Aware Synapses (MAS) \cite{aljundi2018memory} in continual learning, we first design a Na\"ive Implicit-Memory (N-IM) to avoid drastic changes to important parameters learned on old tasks when learning a new task.
	By introducing a regularization term to constrain the parameter update of different sub-modules (in MAttNet or CM-Att-Erase), N-IM can penalize the changes to important parameters, effectively preventing important knowledge related to previous tasks from being overwritten.
	MAttNet and CM-Att-Erase attentively divide the model into different sub-modules related to the subject (e.g., “boy”), the locations (e.g., “in the middle”) and the relationships (e.g., “riding”).
	Considering the sub modular information contained in these individual sub-modules, we further develop a Weighted Implicit-Memory (W-IM) to adaptively adjust the contribution of different sub-modules. 
	By assigning different sub-modules with {task-specific} importance weights, W-IM ensures that those sub-modules that are \textit{sensitive} to the current task are restricted from updates.

	We also develop two effective variants for the Explicit-Memory module, including a Na\"ive Explicit-Memory (N-EM) and a Modular Explicit-Memory (M-EM).
	N-EM selects representative samples of each task for rehearsal by considering the easy ones leading to smaller loss, while M-EM takes into account  not  only  the  easiness of samples  
	but also the importance of the subject-information for individual samples.
	M-EM can quickly retain the knowledge of previous tasks by training the model on rehearsed samples when learning a new task.
	Benefiting from the two modules, i.e., Implicit-Memory and Explicit-Memory, DMM can effectively alleviate the phenomenon of catastrophic forgetting and  continually  improve the grounding performance on incoming tasks, as shown in Fig. \ref{fig:catforgetting} (b).
	
	In summary, our main contributions are three-folds:
	\begin{itemize}
		\item We propose a new, practical and challenging task Continual Referring Expression Comprehension (CREC), which considers a setting in which  training  data  arrives  in  a  streaming fashion.  
		CREC shows better practicality for real-world scenarios. 
		\item We design a baseline method dubbed Dual Modular Memorization (DMM) for CREC. DMM can effectively alleviate the problem of catastrophic-forgetting in CREC with the developed memorization modules of {Implicit-Memory} and {Explicit-Memory}.
		\item We propose three benchmarks for CREC by respectively re-splitting three REC datasets RefCOCO, RefCOCO+ and RefCOCOg into sequential tasks. 
		Extensive experiments on the three benchmarks show that DMM significantly outperforms other alternatives based on two popular REC  backbones.
		We make the source code and benchmarks public available to encourage future research in this field.
	\end{itemize}

	\section{Related work}
	In this section, we review previous research closely related to our method, specifically including  continual learning and referring expression comprehension (REC).
	
	\subsection{Continual Learning} 
	Continual learning is a practical learning mechanism in which a model learns from a stream of incoming data. As the model is updated continually using new tasks, the key challenge in continual learning is to overcome {catastrophic-forgetting}, i.e., how to prevent the model from forgetting previously learned tasks. Catastrophic-forgetting occurs when the model parameters obtained from training on task $A$ change when training on task $B$, which can easily lead to a sharp drop in the results on task $A$. Approaches to addressing the catastrophic forgetting problem can be grouped into three broad categories: regularization-based approaches~\cite{aljundi2018memory,zenke2017continual}, architecture growth approaches\cite{rusu2016progressive,veniat2020efficient}, and replay/rehearsal based approaches \cite{lopez2017gradient,sprechmann2018memory}.

	Regularization-based methods aims at avoiding excessive changes in the parameters learned on old tasks when learning a new task, thus ensuring the accuracy of the network on the old tasks~\cite{aljundi2018memory,zenke2017continual}. 
	Typically, these methods estimate importance weights for each model parameter, and the changes of the important parameters are penalized by a regularizer for previous tasks.
	The dynamic growth approaches methods directly add or modify the model structure of REC. \cite{rusu2016progressive} adds an additional network to each task and lateral connections to the network of the previous task. \cite{veniat2020efficient} proposes a modular layer network approach, whose modules represent atomic skills that can be composed to perform a certain task, and provides a learning algorithm to search the modules to combine with.
	For replay-based methods, catastrophic-forgetting is avoided by storing data from previous tasks and training them together with data from the current task. \cite{lopez2017gradient,sprechmann2018memory,rebuffi2017icarl} use replayed samples from previous tasks to constrain the parameters' update when learning the new task. Most recently, \cite{ayub2021eec} creates reconstructed images from encoded episodes and dynamically generates pseudo-images for model-optimization. 
	Recently, a series of continual learning methods combined with meta-learning have been proposed.
	\cite{DBLP:conf/iclr/RiemerCALRTT19} uses the fixed loss function to align the gradients and updates the network that is well-aligned.
	\cite{DBLP:conf/nips/JavedW19} proposes an approach to disentangle generic representations by task-specific learning.
	
	While the problem of continual learning has been traditionally addressed in image classification, much less attention has been devoted to REC.
	Here we fill this gap, proposing to use the existing continual learning method to solve the REC work. 
	However, it treats the parameters as isolated elements which may result in ignoring the modular information.
	While in CREC, the information involved in sub-modules can boost the learning performance.
	Our designed CREC method, Dual Modular Memorization (DMM), can effectively investigate the modular information to alleviate the problem of catastrophic-forgetting.
	
	\subsection{Referring Expression Comprehension} 
	With the increasing interest in human-computer communication, REC has achieved  great success in recent years. Early models \cite{yu2016modeling,mao2016generation,hu2016natural} use the CNN-LSTM architecture to predict the referent within an image. These networks use deep CNNs to extract features and LSTMs to match the extracted features with the word-vector of the expression to find the referred region. 
	Another line of works \cite{chen2017query,luo2017comprehension,rohrbach2016grounding} treat REC as a cross-domain matching problem, where the expression feature and region feature are embedded into a shared space to measure their compatibility. 
	Recently, \cite{wang2019neighbourhood,yang2019cross,yang2019dynamic} introduce graph technologies to explore the topology structure of images by modeling objects as graph nodes.
	\cite{DBLP:conf/iclr/MaoGKTW19} use curriculum learning to guide the searching over the large compositional space of images and language. \cite{DBLP:conf/cvpr/ZengXHCTG20}  propose a simple but effective IoU regression head module to explicitly consider the localization quality of the grounding results.
	
	While these results are impressive, those REC approaches employ a two-stage learning process. Firstly, an external target detector such as Faster RCNN \cite{7485869} is used to recognize the input image and generate a series of object proposals. Then computing the matching score between these object proposals and the given referring expression, and selecting the target region with the highest matching score as the final results. The limitations are obvious for those two-stage approaches. On the one hand, the use of an external target detector demands additional computational effort. On the other hand, the quality of the object proposals extracted by the target detector affects the performance. 
	To conquer these issues, one-stage REC \cite{Yang_2019_ICCV,yang2020improving,liao2020real,DBLP:conf/cvpr/ZengXHCTG20} has been proposed to process the original images and referring expressions in an end-to-end learning manner.
	
	Significantly, the focus of all those REC works lies in how to more effectively model the language and image to achieve a better REC performance in a stationary evaluation setting. 
	That is, all the object categories are known in advance. However, the focus of our work is orthogonal to such works in that we aim to propose a new REC framework that can work under a continual setting, where the object categories emerge sequentially.
	
	\section{Method}
	In this section, we first describe our proposed Continual  Referring  Expression Comprehension (CREC) task in detail. Then we introduce our baseline Dual Modular Memorization (DMM), specifically including its key components Implicit-Memory and Explicit-Memory.
	
	\subsection{Problem Formulation and Background}
	\label{Formulation}
	Given a referring expression $r$ and an image $I$, the goal of the REC is to localize the object $o$ being referred to by the expression $r$ within the image, by predicting its bounding box $y$. Additionally, each object also belongs to a category $c\in\mathcal{C}$ from the set of categories $\mathcal{C}$. Thus, more formally, each sample in REC can be represented as a tuple $(r,y,c,o)$.
	
	MAttNet \cite{yu2018mattnet} considers the complex linguistic and visual structures by decomposing the expression into three different modular components and designing visual features for each module accordingly. Specifically, given a sample that contains a referring expression $r$ and an image $I$ with a set of {object candidates ${o_i}$}. The model is trained to predict the object with the highest probability in the image. Given an expression $r$, a self-attention mechanism is used to softly decompose it into three modular components, i.e., $subject$, $location$, and $relation$. The final matching score is calculated as the weighted sum of the three matching scores obtained from these three modules. 
	CM-Att-Erase \cite{liu2019improving} is a recently proposed strong baseline in REC, which is built on MAttNet for encouraging the model to explore complementary cross-modal alignments. 
	
	Our designed DMM method adopts the modular structure of MattNet/CM-Att-Erase and makes some important modifications to avoid catastrophic-forgetting, which will be illustrate in Section \ref{DMM}.
	
	\subsection{Continual Referring Expression Comprehension}
	\label{sectionTask}
	We  propose Continual Referring Expression Comprehension (CREC) to improve the practicality  of  REC for  real-world  scenarios. Different from  the  standard  REC which  trains the  model in a one-step  way,  CREC  considers a continual learning setting  where the  training data  arrives in a streaming fashion, as depicted in Fig. \ref{fig:REC_setting}. 
	
	\subsubsection{Task Construction}
	Since CREC is designed to solve the REC problem under the continual learning setting, 
	the critical step is to construct new benchmarks that consist a sequence of tasks for CREC.
	We create three CREC datasets, CRefCOCO, CRefCOCO+ and CRefCOCOg, by respectively re-splitting the three standard REC datasets RefCOCO, RefCOCO+~\cite{yu2016modeling} and RefCOCOg~\cite{mao2016generation} into sequential tasks. 
	In specific, two task sequences with different lengths (5 and 10) were created based on the object super-categories. Here we denote the constructed task sequences as $\mathcal{T} = \{\mathcal{T}_1,\mathcal{T}_2,\ldots,\mathcal{T}_N\}$, where $N$ is the total number of tasks. More details are given in Section \ref{datasplit}.
	
	\subsubsection{Training Strategy}
	In CREC, we denote the data in task $\mathcal{T}_t$ by $D_t = \{(r_i,y_i,c_i,o_i)_{i=1}^{M_t}\}$, where {$(r_i,y_i,c_i,o_i)$} is the \textit{i}-th training sample $x_i$ in task $\mathcal{T}_t$, and $M_t$ denote the total number of samples in the task $\mathcal{T}_t$. 
	Different tasks have no overlap, i.e., $\forall$ $i, j$ and $i \ne j$, $D_i \cap D_j = \oslash$. The tasks $\mathcal{T} = \{\mathcal{T}_1,\mathcal{T}_2,\ldots,\mathcal{T}_N\}$ are revealed to the model sequentially. When task $\mathcal{T}_t$ is presented, the proposed CREC model is trained on $\mathcal{T}_t$ with $D_t$ before the next task $\mathcal{T}_{i+1}$ arrives. In addition, model cannot access the samples from previous tasks $\{\mathcal{T}_1,\mathcal{T}_2,\ldots,\mathcal{T}_{t-1}\}$. At inference, without being given the object category $C_i$, the model needs to ground the object referred to by $r_i$ from image $I_i$ to generate $\hat{y}_i$, if the IOU of $\frac{\hat{y}_i}{y_i}$ is more than threshold, the ground is correct. As the threshold setting, we use the conventional REC setting \cite{liu2019improving}.

	\subsection{Dual Modular Memorization}
	\label{DMM}
	To address the problem of catastrophic forgetting, we develop a novel Dual Modular Memorization (DMM) mechanism for CREC, which consists of two modules, Implicit-Memory and Explicit-Memory, as described in Fig. \ref{fig:DMM}.

	\begin{algorithm}[t]
		\caption{Modular Explicit-Memory for Rehearsal.}
		\label{algorithm1}
		{\bf Input:}
		Model $F$; Training task $\mathcal{T}_t$; Task-size for the current task $D_t$; Task-size for the previous tasks $D={D_1,...,D_{t-1}}$; Max memory-size $K$; Buffer pool $B={B_1,...,B_{t-1}}$. \\
		{\bf Output:}
		Optimized $F$ and updated $B$, $D$.
		\begin{algorithmic}[1]
			\For {\{${x,y}\} \in D_t$,\{${x_B,y_B}\} \in B$} \algorithmiccomment{joint training.}
			\State ${\{\hat{y}\}}, Att_{sub} =F(\theta,{x})$
			\State $LLoss= L(\{\hat{y}\},\{y\})$
			\State $Hloss=L(F,\{x_B,y_B\})$ 
			\State $F\longleftarrow Backprob(LLoss+Hloss)$
			\EndFor
			\State $D_{sum}= {\textstyle \sum_{i=1}^{t}}D_i $
			\For {$i=0,\ldots,t-1$} \algorithmiccomment{reorganize the buffer pool.}
			\State $D_i= K \cdot \frac{D_i}{D_{sum}} $
			\State $B_i \longleftarrow SortBufferPool(B_i,D_i)$ %
			\EndFor
			\State $D_t= K \cdot \frac{D_t}{D_{sum}}$
			
			\For {\{${x,y}\} \in D_t$} \algorithmiccomment{update the memory.}
			\State ${\{\hat{y}\},Att_{sub}} =F(\theta,{x})$
			\State $Loss=Att_{sub} \cdot L(\{\hat{y}\},\{y\})$
			\State $B_t \longleftarrow \{x,y\}$
			\EndFor
			
			\State $B \longleftarrow \{B_1,...B_t\},D \longleftarrow \{D_1,...D_t\}$
			\State \Return $F,B,D$
		\end{algorithmic}
	\end{algorithm}

	\subsubsection{Implicit-Memory}
	\label{IM}
	Before delving into our implicit-memory, we introduce some notations as follows. Let $F^t$ represent our model DMM for each task $\mathcal{T}_t$. We split $F^t$ into three parts, i.e., $F^t_{sub},F^t_{loc}$ and $F^t_{rel}$. The $F^t_{sub},F^t_{loc}$ and $F^t_{rel}$ denote subject module, location module and relation module, respectively.
	
	\textbf{Na\"ive Implicit-Memory (N-IM):}
	\label{N-IM}
	To guarantee the stability of model, an intuitive idea to alleviate the catastrophic forgetting is to use the regularization to constrain the parameters updating. This helps to remember the important parameters learned on the previous tasks when learning a new task. 
	To achieve this, we benefit the regularization from MAS \cite{aljundi2018memory} to CREC, which can be formulated as:
	\begin{align}
		\label{align1}
		g^t_{i}(x_k) &=\frac{\partial (F^t_{}(x_k;\theta_{i} ))}{\partial \theta_{i}},\\
		\label{align2}
		\Omega^t_{i} &=\frac{1}{M_t}{\textstyle\sum_{k=1}^{M_t}}\left\|g^t_{i}\left(x_{k}\right )\right\|,
	\end{align}
	where $M_t$ is the number of samples in the task $\mathcal{T}_t$; $g^t_{i}(x_k)$ is the gradient of the loss of $x_k$ in $\mathcal{T}_t$, w.r.t.\ the $i$-th parameter $\theta_{i}$. We accumulate the gradients $g^t_{i}(x_k)$ to obtain the importance weight $\Omega^t_{i}$, by Eq. \ref{align2}.
	In order to calculate the loss $L(\theta )$ when learning on the new task $\mathcal{T}_{t+1}$, we add a regularization term to $L_{t+1} (\theta)$ which calculated by tradition REC training for avoiding  drastic changes to the important parameters:
	\begin{equation}
		\label{equation2}
		L(\theta )=L_{t+1} (\theta )+\frac{\lambda }{2} {\sum}_{i}\Omega^t _{i}(\theta _{i}-\theta _{i}^{\ast } )^2.
	\end{equation}
	where $\lambda$ is a hyper-parameter to balance the loss of the new task and the parameter change constraint; $\theta_{i}^*$ denotes the optimal parameter for the previous task. 
	
	However, a critical limitation for na\"ive implicit-memory is that it treats the model parameters as isolated elements and ignores the complex structural information contained in individual sub-modules. 
	Thus, before determining the parameter-task association, we need to consider in advance which sub-module should be retained for a task. If a module is widely shared among tasks, it makes less sense to memorize the parameters of this module.

	\textbf{Weighted Implicit-Memory (W-IM):}
	\label{W-IM}
	To better estimate the importance of sub-modules and the parameters within each sub-module, we propose the weighted implicit-memory, where importance of sub-module is introduced to constrain the updating.
	Specifically, the objective of weighted implicit-memory is defined as:
	\begin{equation}
		\Omega^t_m = \frac{1}{\left|F_m\right|\cdot M_t}{\sum}_{g^t_{i}\in F_m} {\textstyle \sum_{k=1}^{M_t}} \left \| g^t_{i}(x_k) \right \|,
	\end{equation}
	
	\begin{equation}
		W^t_{\Omega_m} =\frac{\Omega^t_m}{ {\textstyle\sum_{m}}\Omega^t}.
	\end{equation}
	where $\left|F_m\right|$ is the total parameter number of $F_m$, $m \in \{sub,loc,rel\}$. The weights of the modules are normalized by the sum of three modules' weights, $W^t_{\Omega_{m}}$ is the normalized importance matrix. A higher module weight indicates that the module is important for learning a task, so the parameters within it should be less updated. 
	Hence, the learning objective is defined as follows:
	\begin{equation}
		L(\theta )=L_{t+1} (\theta )+\frac{\lambda }{2} {\sum}_{m}{\sum}_{i \in F_m}W^t_{\Omega_m}\Omega^t _{i}(\theta _{i}-\theta _{i}^{\ast } )^2.
	\end{equation}
	Note that both $\Omega_{i}$ and $W_{\Omega_{m}}$ are updated by accumulating previous estimations when a new task arrives.
	
	\subsubsection{Explicit-Memory}
	\label{EM}
	By constraining dramatic changes to important parameters of the model, the designed implicit-memory regularization objectives N-IM and W-IM can effectively address the catastrophic forgetting.
	In order to further improve the memory capacity of the model, we further propose an Explicit-Memory module to explicitly store some samples from previous tasks which are representative.
	
	\textbf{Na\"ive Explicit-Memory (N-EM)}:
	\label{N-EM}
	An intuitive idea to select representative samples of a task is to select \emph{easy} samples leading to a small loss 
	--  samples with smaller training losses can be easily used to learn the task-specific knowledge for the corresponding task.
	In this way, we can directly store representative samples of each seen task in a buffer pool for future rehearsal.
	We use different flags to mark the loss calculated based on different samples.
	The “LLoss” represents the loss calculated by samples of current task. 
	And the “HLoss” represents the loss calculated by samples of buffer pool.
	However, this \textit{loss-oriented} strategy ignores the impact of subject intrinsic information on performance.

	\textbf{Modular Explicit-Memory (M-EM):}
	\label{M-EM}
	The Na\"ive Explicit-Memory just takes into account the loss of samples.
	However, it is not good enough.
	So we propose modular explicit-memory (M-EM) by marrying the ideas of modular decomposition and vanilla explicit-memory.
	Considering a simple scenario where a sample $X_i$ and a sample $X_j$ hold the same loss $L$. The sample $X_i$ has more intrinsic information about the current task, while sample $X_j$ has less. We would intuitively choose the sample $X_i$ to save for rehearsal. So the ability to evaluate how much intrinsic information the sample contains about the current task for N-EM is important.
	As mentioned in \ref{Formulation}, the model has a self-attention mechanism which can estimate the weight value of the three modules. So, the weight value can represent the importance of the intrinsic information. In other words, the module with higher weight value contains more intrinsic information about the current task.
	In addition, we re-split the datasets by class of the samples. 
	And in the expression, the subject is the instance of the class.
	An intuitive way is to choose the subject module. 
	The empirical analysis in Section \ref{ablation-submodule} which is conducted on our re-split datasets also confirms our intuition that the sub-module has a bigger weight than other modules, indicating that the sub-module is the most important module. 
	Therefore, we regard the sub-module weight is helpful for sample selection.
    So this modular version takes into account not only the loss but also how importance of the subject's intrinsic information in each sample.
	We implement it by directly multiplying the subject-weight of the self-attention $Att_{sub}$ to the loss of the sample.
	
	\begin{table}[t]
		\renewcommand\arraystretch{1.3}
		\renewcommand\tabcolsep{4.0pt}
		\caption{Brief statistics of the constructed 5-task datasets for CREC. The entries show the number of samples for each task in each dataset.}
		\label{tab:datatable}
		\centering
		\begin{tabular}{llrrrrr}
			\toprule
			Dataset & split & task1 & task2 & task3 & task4 & task5 \\ \midrule
			CRefCOCO & train & 60357 & 16748 & 16087 & 14079 & 13353 \\
			& val & 5498 & 1580 & 1318 & 1417 & 1021 \\
			& test & 5273 & 1707 & 1307 & 1314 & 1151 \\ \midrule
			CRefCOCO+ & train & 61292 & 15888 & 15957 & 13914 & 13230 \\
			& val & 5568 & 1483 & 1298 & 1385 & 1024 \\
			& test & 5430 & 1577 & 1300 & 1277 & 1121 \\ \midrule
			CRefCOCOg & train & 30712 & 10533 & 12105 & 13526 & 13636 \\
			& val & 1776 & 673 & 743 & 866 & 838 \\
			& test & 3477 & 1198 & 1485 & 1747 & 1695 \\ \bottomrule
		\end{tabular}
	\end{table}
	
	\begin{table*}[t]
		\renewcommand\arraystretch{1.3}
		\renewcommand\tabcolsep{4.0pt}
		\caption{
			Quantitative performance comparison of different state-of-the-art methods with the DMM on our re-split datasets under the 5-task setting. Best scores among all methods are in bold. LA, AA, FWT and BWT is the last accuracy, average accuracy, forward transfer and backward transfer, respectively. Higher is better for all metrics.
		}
		\label{tab:result}
		\centering
		\begin{tabular}{lccccccccccccc}
			\toprule
			&  & \multicolumn{4}{c}{CRefCOCO} & \multicolumn{4}{c}{CRefCOCO+} & \multicolumn{4}{c}{CRefCOCOg} \\
			Method & Backbone & LA & AA & FWT & BWT & LA & AA & FWT & BWT & LA & AA & FWT & BWT \\ \midrule
			Joint Training & MAttNet & 85.36 & - & - & - & 71.26 & - & - & - & 78.12 & - & - & - \\
			Finetuning & MAttNet & 53.00 & 69.06 & 41.93 & -26.51 & 39.35 & 57.28 & 18.28 & -23.22 & 52.90 & 63.71 & 24.13 & -26.07 \\
			MAS\cite{aljundi2018memory} & MAttNet & 66.70 & 76.77 & 40.59 & -13.35 & 52.85 & 64.85 & 20.47 & -11.90 & 57.97 & 66.91 & 26.18 & -17.43 \\
			GDumb\cite{prabhu2020gdumb} & MAttNet & 55.55 & 60.37 & 38.00 & -13.43 & 35.93 & 39.76 & 15.59 & -15.49 & 44.10 & 48.88 & 21.73 & -20.80 \\
			EWC\cite{kirkpatrick2017overcoming} & MAttNet & 58.07 & 68.80 & 40.83 & -24.24 & 43.98 & 56.88 & 21.04 & -20.71 & 51.94 & 63.48 & 24.19 & -22.46 \\
			DMM & MAttNet & \textbf{76.12} & \textbf{82.20} & \textbf{43.46} & \textbf{-5.37} & \textbf{62.30} & \textbf{69.38} & \textbf{24.35} & \textbf{-3.93} & \textbf{69.24} & \textbf{75.46} & \textbf{29.92} & \textbf{-4.24} \\ \midrule
			Joint Training & CM-Att-Erase & 86.44 & - & - & - & 72.03 & - & - & - & 80.37 & - & - & - \\
			Finetuning & CM-Att-Erase & 64.46 & 76.62 & 49.47 & -18.54 & 40.34 & 48.51 & \textbf{26.85} & -26.46 & 55.15 & 52.82 & \textbf{22.45} & -26.79 \\
			MAS\cite{aljundi2018memory} & CM-Att-Erase & 74.89 & 82.52 & 45.50 & -8.16 & 50.69 & 58.12 & 25.59 & -14.62 & 64.39 & 68.32 & 18.22 & -8.97 \\
			GDumb\cite{prabhu2020gdumb} & CM-Att-Erase & 20.71 & 30.99 & 5.97 & -18.49 & 19.05 & 22.62 & -0.27 & -16.30 & 20.39 & 14.99 & -9.61 & 7.37 \\
			EWC\cite{kirkpatrick2017overcoming} & CM-Att-Erase & 39.81 & 50.80 & 22.23 & -7.19 & 29.33 & 36.05 & 8.98 & -9.63 & 23.29 & 19.89 & -5.40 & 14.35 \\
			DMM & CM-Att-Erase & \textbf{78.31} & \textbf{83.82} & \textbf{50.02} & \textbf{-5.88} & \textbf{59.52} & \textbf{64.14} & 25.64 & \textbf{-6.12} & \textbf{65.37} & \textbf{69.88} & 17.08 & \textbf{-6.93} \\ \bottomrule
		\end{tabular}
	\end{table*}
	
	Algorithm \ref{algorithm1} describes the details of our modular explicit-memory strategy for rehearsal. 
	Considering the memory efficiency, explicit-memory ensures that the total number of exemplar images never exceeds a fixed parameter $K$ throughout the training stage.
	When the system receives task $\mathcal{T}_t$, the model is jointly updated using the samples from both the current task and buffer pool (lines 1--6).
	After the training process, we compute the percentage of all seen tasks and multiply $K$ to get the memory buffer size of each task. 
	Then the procedure ``SortBufferPool'' on line 10 drops the higher-loss samples to reach the available size for each task (lines 7--11). 
	In the end, we choose the lowest loss samples of the current task $\mathcal{T}_t$ into the memory buffer (lines 12--17).
	Note that in line 15, the loss of a sample $(x,y)$ in $\mathcal{T}_t$ is weighted by the subject-weight $Att_{sub}$, which indicates the importance of the subject module for the sample. 
	Larger $Att_{sub}$ means the language query contains more intrinsic information and less transferable clues.
	We tend to store such samples in the buffer pool so that the model can recall the previous knowledge when it encounters some new information by rehearsing such important samples.

	\section{Experiments}
	\subsection{Dataset Construction}
	
	We create three CREC benchmarks by respectively re-splitting the three standard REC datasets RefCOCO, RefCOCO+~\cite{yu2016modeling} and RefCOCOg~\cite{mao2016generation} into sequential tasks. 
	Correspondingly, we term these three new datasets as CRefCOCO, CRefCOCO+ and CRefCOCOg.
	
	\subsubsection{REC Datasets} 
	The three REC benchmark datasets RefCOCO, RefCOCO+ and RefCOCOg are all constructed from  MSCOCO~\cite{DBLP:journals/corr/LinMBHPRDZ14}. 
	Each dataset contains 80 object categories (e.g.,\ bird, dog, apple and sandwich) and 13 object supercategories (e.g.,\ animal, food). Several characteristics of these datasets are worth mentioning: 
	(1) The average length of textual expressions in RefCOCO and RefCOCO+ are 3.61 and 3.64 words respectively, whereas the expressions in RefCOCOg are longer and more complex, with 8.4 words on average. 
	(2) The images in RefCOCO and RefCOCO+ contain more instances of the same category and thus render more distracting information to localize the referent. 
	(3) Any words depicting absolute locations are forbidden in RefCOCO+, as it focuses on appearance clues. 
	Both RefCOCO and RefCOCO+ are split into the subsets of train, validation, Test A and Test B. The Test A split mainly contains images of ``people'' supercategory. The Test B split contains multiple instances of all the other object supercategories. 
	To better evaluate the model, we combine Test A and Test B to get the Test split. 
	RefCOCOg is split into the subsets of train, validation and test, and the categories are distributed more evenly.
	
	\begin{table*}[tb]
		\renewcommand\arraystretch{1.3}
		\renewcommand\tabcolsep{4.0pt}
		\caption{
			Overall performance comparison with different state-of-the-art methods on our re-split datasets under the 10-task setting. Training settings are the same as Table \ref{tab:result}. Best scores among all methods are in bold.}
		\centering
		\begin{tabular}{lccccccccccccc}
			\toprule
			&  & \multicolumn{4}{c}{CRefCOCO} & \multicolumn{4}{c}{CRefCOCO+} & \multicolumn{4}{c}{CRefCOCOg} \\
			Method & Backbone & LA & AA & FWT & BWT & LA & AA & FWT & BWT & LA & AA & FWT & BWT \\ \midrule
			Joint Training & MAttNet & 85.36 & - & - & - & 71.26 & - & - & - & 78.12 & - & - & - \\
			Finetuning & MAttNet & 55.01 & 72.99 & 37.67 & -11.10 & 39.49 & 50.07 & 18.28 & -10.38 & 49.86 & 60.91 & 22.08 & -14.91 \\
			MAS\cite{aljundi2018memory} & MAttNet & 63.26 & 74.04 & 36.03 & -6.92 & 45.37 & 50.90 & 16.62 & -6.35 & 50.34 & 61.15 & 19.61 & -8.64 \\
			GDumb\cite{prabhu2020gdumb} & MAttNet & 59.40 & 64.28 & 36.93 & -7.44 & 32.07 & 37.05 & 13.00 & -8.03 & 46.81 & 50.89 & 18.64 & -7.80 \\
			EWC\cite{kirkpatrick2017overcoming} & MAttNet & 61.11 & 70.77 & 40.20 & -7.13 & 38.70 & 48.21 & 18.84 & -6.13 & 47.39 & 60.40 & 22.33 & -9.20 \\
			DMM & MAttNet & \textbf{74.76} & \textbf{75.63} & \textbf{41.28} & \textbf{-3.03} & \textbf{57.52} & \textbf{58.53} & \textbf{24.11} & \textbf{-2.97} & \textbf{66.47} & \textbf{64.25} & \textbf{27.08} & \textbf{-2.38} \\ \midrule
			Joint Training & CM-Att-Erase & 86.44 & - & - & - & 72.03 & - & - & - & 80.37 & - & - & - \\
			Finetuning & CM-Att-Erase & 61.79 & 74.11 & 40.59 & -13.67 & 40.12 & 50.85 & 24.58 & -13.00 & 48.70 & 61.62 & 14.72 & -21.29 \\
			MAS\cite{aljundi2018memory} & CM-Att-Erase & 66.93 & 75.16 & 36.49 & -5.13 & 49.34 & 53.54 & 20.74 & -9.43 & 58.05 & 61.90 & 9.52 & -9.51 \\
			GDumb\cite{prabhu2020gdumb} & CM-Att-Erase & 23.73 & 15.08 & -3.94 & \textbf{1.47} & 18.30 & 15.62 & 0.47 & \textbf{-0.71} & 18.59 & 12.84 & -9.39 & \textbf{5.39} \\
			EWC\cite{kirkpatrick2017overcoming} & CM-Att-Erase & 32.70 & 30.95 & 11.00 & 0.94 & 16.11 & 13.16 & -0.40 & -1.54 & 11.16 & 8.93 & -10.27 & 3.05 \\
			DMM & CM-Att-Erase & \textbf{71.37} & \textbf{78.19} & \textbf{46.12} & -4.21 & \textbf{53.72} & \textbf{55.45} & \textbf{26.35} & -1.70 & \textbf{62.12} & \textbf{65.10} & \textbf{17.42} & -4.24 \\ \bottomrule
		\end{tabular}
		\label{tab:10task_result}
	\end{table*} 
	
	\subsubsection{CREC Benchmarks}
	\label{datasplit}
	In order to evaluate the continual learning capability of our DMM method, 
	we separately re-split RefCOCO, RefCOCO+~\cite{yu2016modeling} and RefCOCOg  into subsets/tasks according to the object supercategories of each dataset. 
	Specifically, two task sequences with different lengths are created.
	
	\textbf{10-task}:
	Following our problem formulation in Section~\ref{Formulation}, we treat an object supercategory as a task. 
	We sort the supercategories by their number of samples. The 10 most frequent supercategories are adopted as disjoint tasks, specifically including: $Food$, $Indoor$, $Sports$, $Person$, $Animal$, $Vehicle$, $Furniture$, $Accessory$, $Electronic$, and $Kitchen$.
	
	\textbf{5-task}:
	As some supercategories have too few samples, e.g.,\ 853 samples in the $Sports$ supercategory, we merge several supercategories to form a new supercategory according to the similarity between them. 
	In this way, 5 disjoint tasks with balanced sample numbers are constructed, including Task1: {$Person$}; Task2: $Kitchen+Food$; Task3: $Animal$; Task4: $Indoor+Appliance+Furniture+Electronic$; Task5: $Outdoor+Vehicle+Sports+Accessory$.
	The number of referring expressions per supercategory of the five tasks is shown in Table~\ref{tab:datatable}.

	\begin{table*}[t]
		\renewcommand\arraystretch{1.3}
		\renewcommand\tabcolsep{4.0pt}
		\caption{
			Quantitative performance evaluation of different components in DMM under 5-task with MAttNet for the ablation study, training sequence and settings are the same as table \ref{tab:result}. The first row is the Finetuning. N-IM, W-IM, N-EM and M-EM denote na\"ive implicit-memory, weighted implicit-memory, na\"ive explicit-memory and modular explicit-memory. Best scores among all methods are in bold.}
		\label{tab:ablationtab}
		\centering
		\begin{tabular}{clllcccccccccccc}
			\hline
			\multicolumn{4}{c}{Component} & \multicolumn{4}{c}{CRefCOCO} & \multicolumn{4}{c}{CRefCOCO+} & \multicolumn{4}{c}{CRefCOCOg} \\ \hline
			N-IM & W-IM & N-EM & M-EM & LA & AA & FWT & BTW & LA & AA & FWT & BTW & LA & AA & FWT & BTW \\ \hline
			&  &  &  & 53.00 & 69.06 & 41.93 & -26.51 & 39.35 & 57.28 & 18.28 & -23.22 & 52.90 & 63.71 & 24.13 & -26.07 \\
			\Checkmark &  &  &  & 66.70 & 76.77 & 40.59 & -13.35 & 52.85 & 64.85 & 20.47 & -11.90 & 57.97 & 66.91 & 26.18 & -17.43 \\
			& \Checkmark &  &  & 69.89 & 77.97 & 44.26 & -10.92 & 46.90 & 62.09 & 21.65 & -15.48 & 61.86 & 69.53 & 28.96 & -15.20 \\
			&  & \Checkmark &  & 68.46 & 78.77 & 42.44 & -11.37 & 50.78 & 64.91 & 20.56 & -11.81 & 62.79 & 73.25 & 26.92 & -9.64 \\
			&  &  & \Checkmark & 71.98 & 79.08 & \textbf{44.43} & -10.47 & 51.95 & 65.34 & 24.24 & -10.99 & 65.84 & 73.29 & 25.01 & -10.11 \\
			\Checkmark &  & \Checkmark &  & 76.04 & 81.38 & 44.07 & -5.73 & 61.14 & 68.62 & 24.08 & -5.32 & 69.03 & 75.19 & 29.88 & -4.39 \\
			& \Checkmark &  & \Checkmark & \textbf{76.12} & \textbf{82.20} & 43.46 & \textbf{-5.37} & \textbf{62.30} & \textbf{69.38} & \textbf{24.35} & \textbf{-3.93} & \textbf{69.24} & \textbf{75.46} & \textbf{29.92} & \textbf{-4.24} \\ \hline
		\end{tabular}
	\end{table*}
	
	\subsection{Experiment Setup}
	\subsubsection{Evaluation Metric} 
	\label{Evaluation Metric}
	Given an input sample $(r,y,c,o)$, let $\hat{y}$ represent the bounding box predicted by a model for expression $r$. We employed Intersection over Union (IoU) as a basic metric to determine
	whether a comprehension is positive or not.
	The IOU is defined as:
	\begin{equation}
		IOU = \frac{\textit{intersection}(y,\hat{y} )}{union(y,\hat{y} )}. 
	\end{equation}
	
	Following the standard setting \cite{liu2019improving}, if the IOU score is greater than threshold 0.5, 
	we consider the prediction to be correct. 
	We evaluate the performance of alleviating catastrophic forgetting on the following metrics:
	Last Accuracy (LA), Average Accuracy (AA), Forward Transfer (FWT) and Backward Transfer (BWT) ~\cite{lopez2017gradient}. 
	
	\textbf{i)} LA is the final accuracy result on the whole test set at the end of training on all tasks in the continual learning process. 
	
	\textbf{ii)} AA evaluates model performance on all tasks seen up to step $i$ once the model is trained on task $\mathcal{T}_i$ by
	\begin{equation}
		\text{AA} = \frac{1}{i}{\textstyle \sum_{j=1}^{i}a_{i,j}}.
	\end{equation}
	where $a_{i,j}$ is the accuracy evaluated on the test set of task $j$ after training the model from task 1 through to $i$.
	
	\textbf{iii)} FWT measures a model's capability of tranferring konwledge from past tasks when learning a new task. 
	Concretely, after training on $\mathcal{T}_i$, we evaluate the model on unseen tasks $T_j \in \{\mathcal{T}_{i+1},\ldots,\mathcal{T}_N\}$ by:
	\begin{equation}
		\text{FWT} = \frac{1}{N-i}  {\textstyle \sum_{j=i+1}^{N}}(a_{i,j}-b_j).
	\end{equation}
	where $b_i$ represent the test accuracy on task $T_i$ with  random initialization.
	
	\textbf{iv)} BWT measures the model's capability of retaining previous knowledge after learning a new task. That is, after training on $\mathcal{T}_i$, the model can be evaluated on $j \in \{\mathcal{T}_1,\ldots,\mathcal{T}_{i-1}\}$ by:
	\begin{equation}
		\text{BWT} = \frac{1}{i-1} {\textstyle \sum_{j=1}^{i-1}} (a_{i,j}-a_{j,j}).
	\end{equation}
	
	For each evaluation metric, we take the average of result of each task as the final evaluation result.
	The larger these metrics, the better is the model. 
	Obviously, it is meaningless to compute the FWT for the first task and the BTW for the last task~\cite{lopez2017gradient}.
	
	\begin{figure*}[t]
	\centering
	\includegraphics[width=\textwidth]{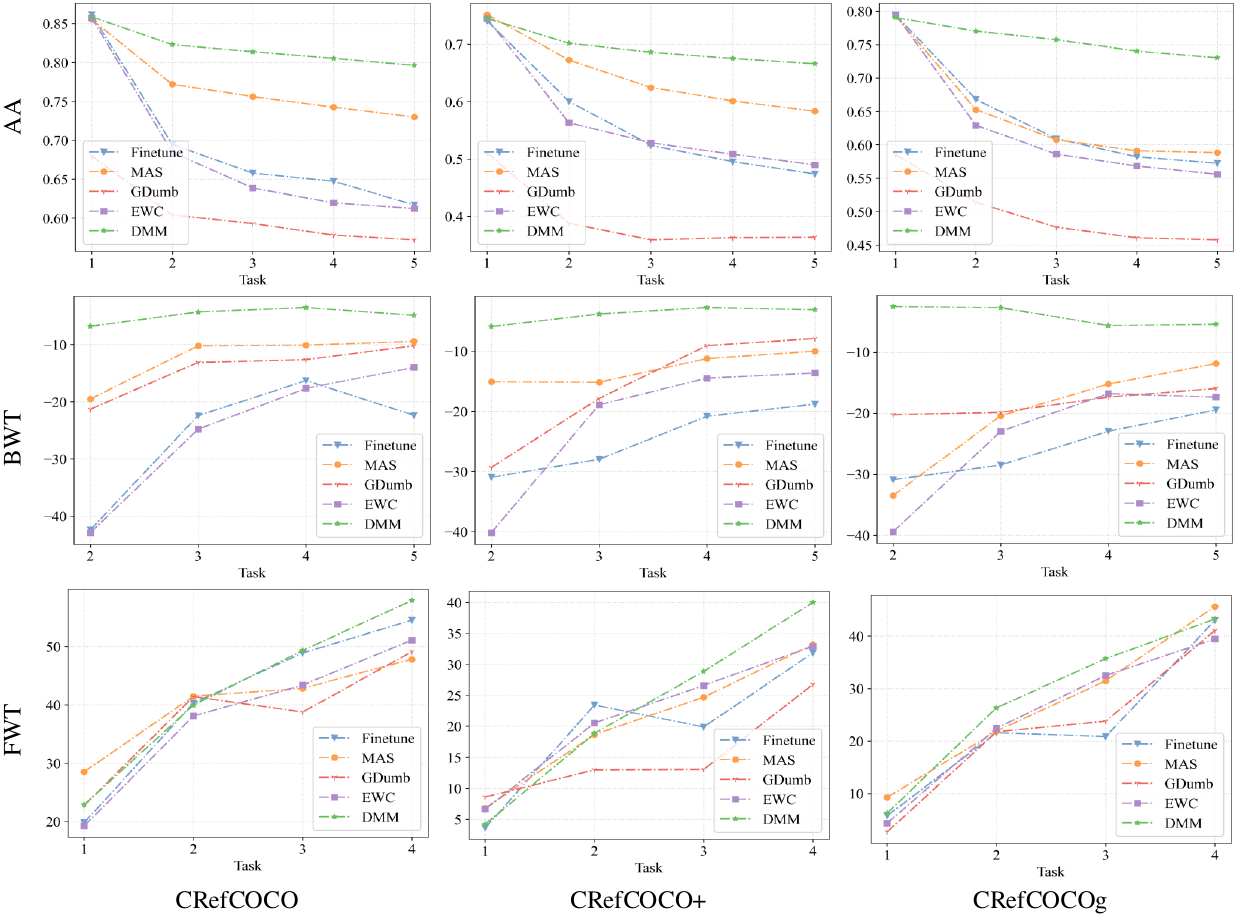}
	\caption{Performance comparison with different state-of-the-art methods under the 5-task setting w.r.t. (Top) AA, (Middle) BWT and (Bottom) FWT metrics. We use MAttNet as the backbone. Models are evaluated after training on each task.}
	\label{fig:acc_line}
	\end{figure*}

	\subsubsection{Implementation Details} We use MAttNet \cite{yu2018mattnet} and CM-Att-Erase \cite{liu2019improving} as backbone models to validate the effectiveness of our proposed DMM method. Mask R-CNN \cite{He_2017_ICCV} with ResNet-101 \cite{He_2016_CVPR} is used as the backbone to extract visual representations. As the regularization parameter $\lambda$, we use the setting in \cite{aljundi2018memory} and set it to 1 for all experiments. The memory size $K$ in the Explicit-Memory is set to 120. In the 5-task setting, as the number of samples of Task1 is more than other tasks, we train the model for 40 epochs for Task1 and 20 epochs for each of the other tasks. One training batch contains 45 referring expressions. Other settings are the same as in the baseline models. Furthermore, for each task setting and backbone, we conduct two experiments to obtain the average result as the final result. The network is implemented based on PyTorch \cite{DBLP:journals/corr/abs-1912-01703}.
	
		\begin{figure*}[t]
		\centering
		\includegraphics[width=\textwidth]{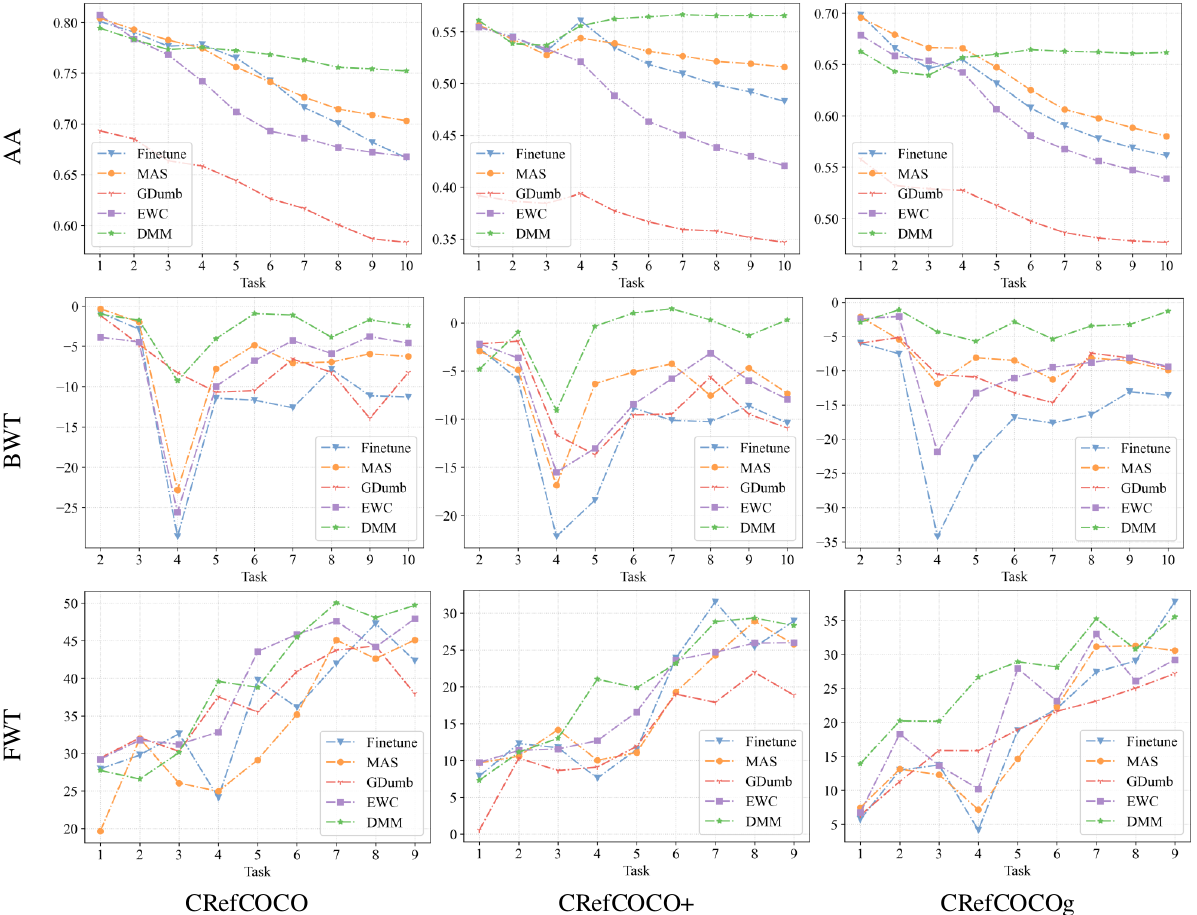}
		\caption{
			Performance comparison of (Top) AA, (Middle) BWT and (Bottom) FWT metrics with different state-of-the-art methods under the 10-task setting. We use MAttNet as the backbone. Models are evaluated after training on each task.}
		\label{fig:acc_line10}
	\end{figure*}
	
	\subsection{Experimental Results}
	
	To evaluate the effectiveness of our proposed DMM method, we first compare it with state-of-the-art continual learning methods under the 5-task and 10-task settings. 
	Then, we conduct ablation studies to further investigate the effectiveness of main components.
	
	\subsubsection{Comparison with State-of-the-arts}
	We compare our method DMM with several state-of-the-art continual learning methods, including Joint Training, Finetuning, MAS \cite{aljundi2018memory}, GDumb~\cite{prabhu2020gdumb} and EWC~\cite{kirkpatrick2017overcoming}. 
	In particular, Joint Training considers all data in the task sequence simultaneously.
	This baseline represents the performance upper bound. 
	Finetuning trains a single model to solve all the tasks without any regularization and initializes from the model of the previous task, i.e., it represents a model trained in the conventional supervised setting.

	\begin{table*}[t]
		\renewcommand\arraystretch{1.3}
		\caption{Quantitative performance comparison of ablation study about samples choosing in modular explicit-memory under 5-task setting. We choose MAttNet as the backbone. Best scores among all methods are in bold.}
		\centering
		\begin{tabular}{ccccccccccccccc}
			\hline
			\multicolumn{3}{c}{Strategy} & \multicolumn{3}{c}{CRefCOCO} &  & \multicolumn{4}{c}{CRefCOCO+} & \multicolumn{4}{c}{CRefCOCOg} \\ \hline
			low & random & high & LA & AA & FWT & BWT & LA & AA & FWT & BWT & LA & AA & FWT & BWT \\ \hline
			\checkmark &  &  & \textbf{71.98} & \textbf{79.08} & \textbf{44.43} & \textbf{-10.47} & \textbf{51.95} & \textbf{65.34} & \textbf{24.24} & \textbf{-10.99} & \textbf{65.84} & \textbf{73.29} & \textbf{25.01} & \textbf{-10.11} \\
			& \checkmark &  & 69.43 & 78.23 & 43.42 & -12.57 & 51.78 & 64.36 & 21.34 & -11.98 & 59.87 & 62.59 & 12.24 & -25.66 \\
			&  & \checkmark & 63.36 & 75.64 & 43.59 & -16.05 & 45.35 & 63.92 & 18.44 & -14.66 & 57.62 & 62.23 & 17.95 & -27.03 \\ \hline
		\end{tabular}
		\label{tab:em loss}
	\end{table*}
	
	\textbf{5-task setting}:
	The results of the experiment under the 5-task setting are shown in Table \ref{tab:result}. 
	As can be observed from the table, compared to Joint Training, Finetuning suffers a significant performances decrease on all three datasets. This shows that Finetuning suffers severe catastrophic forgetting in CREC.
	In addition, the boost in performance brought by MAS is observed on both datasets and on all the accuracy metrics employed.
	As can be seen from the table, DMM consistently outperforms all other methods by a significant margin on the three benchmark datasets in all but two cases. 
	On average, DMM outperforms Finetuning by 17.61\% and 12.81\% in terms of LA and AA, respectively. This clearly shows that our network achieves notable stability and plasticity in CREC.
	
	In order to further compare the different baselines, we plot the results as shown in Fig. \ref{fig:acc_line}.
	As shown in the top panel of Fig.\ \ref{fig:acc_line}, it can be observed that our method consistently surpasses other counterparts at every task on all three datasets on average accuracy.
	In the middle panel of Fig.\ \ref{fig:acc_line}, DMM performs better than other methods on backward transfer (BWT) at each task on all datasets, which indicates that our method guarantees the stability of the network.
	Finally, Fig. \ref{fig:acc_line} (Bottom) shows FWT values for each task in comparison to prior methods for CREC on three datasets. Our results suggest our model achieves notable plasticity.
	
	\begin{figure}[t]
		\centering
		\includegraphics[width=.9\linewidth]{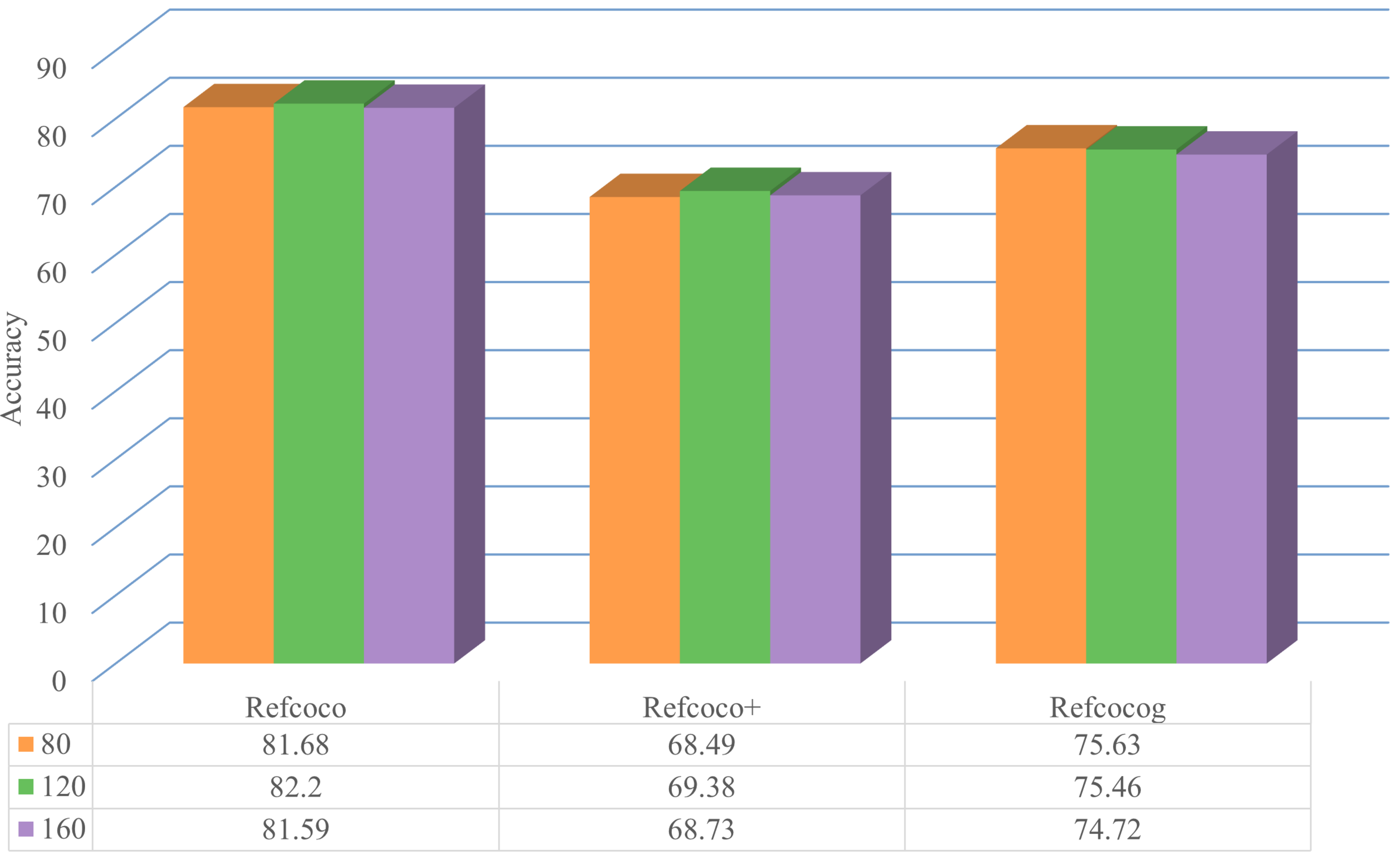}
		\caption{Quantitative results in terms of average accuracy on three different datasets about various memory size, including 80, 120, 160. The results are roughly consistent on all three datasets, showing that our model is not sensitive to the memory size.}
		\label{fig:memorysize}
	\end{figure}

	\textbf{10-task setting}:
	In order to further evaluate the ability to prevent catastrophic forgetting on the longer sequence of the proposed DMM, we conduct the experiments under the 10-task setting. The experimental results are shown in Table \ref{tab:10task_result}.

	As expected, the proposed method is significantly better than almost all other methods under the 10-task setting. These results illustrate the strong ability of our DMM model to alleviate the catastrophic forgetting problem over longer sequences.
	It is worth noting that GDumb achieves better results in terms of BWT with CM-Att-Erase as the backbone. However, GDumb's performance is the worst in terms of LA and AA with both backbones. A possible explanation for GDumb's best BWT performance is that it learns little knowledge when training, so it has nothing to forget.
	In addition, we also plot the results of DMM in comparison with prior methods in Fig.~\ref{fig:acc_line10}. 
	These plots further demonstrate the performance advantages of our method. 
	\begin{figure*}[t]
		\centering
		\includegraphics[width=.9\textwidth]{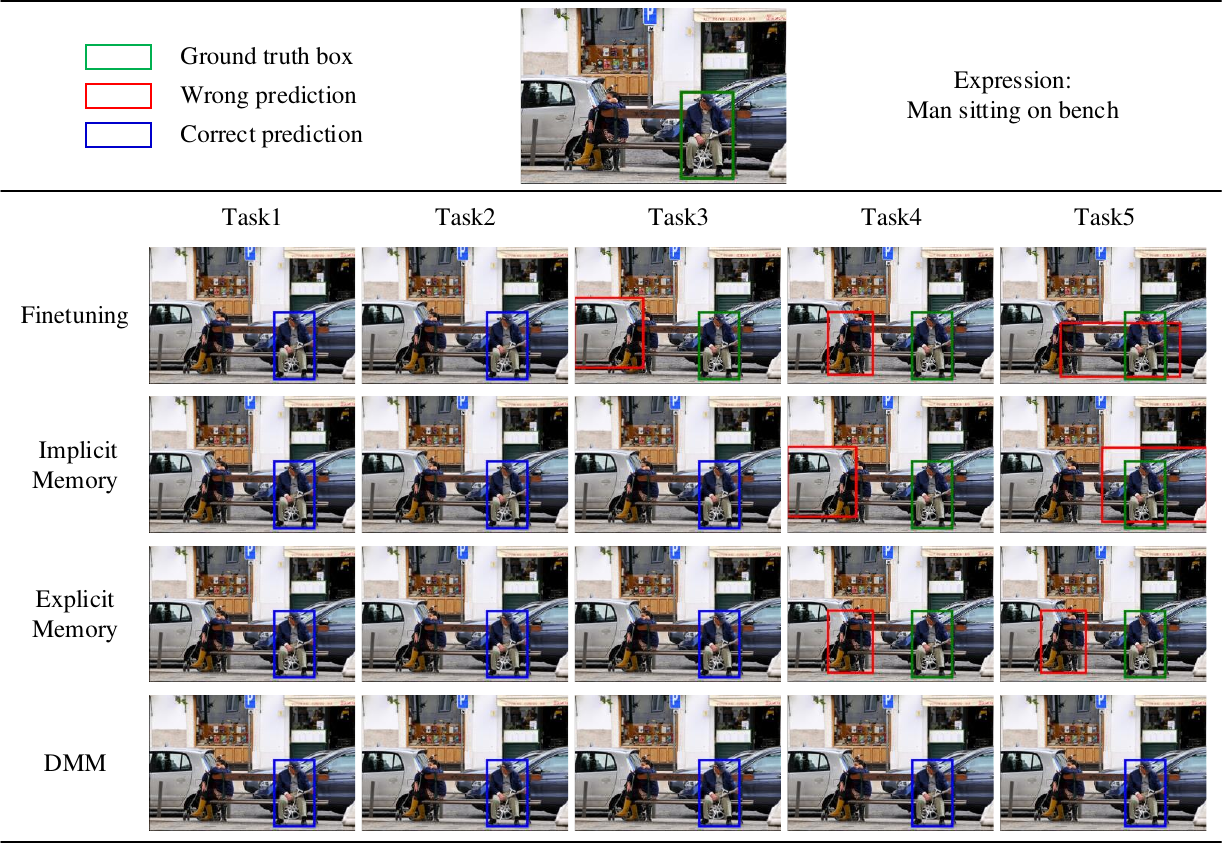}
		\caption{Qualitative evaluation of CREC under 5-task setting. From top to bottom are the ground-truth of the example, the results produced from Finetune, weighted implicit-memory, modular explicit-memory, and the results of Dual Modular Memorization. The five images from left to right of each row denote the localization results after learning the $i$-th task. The example belongs to the Task1.}
		\label{fig:qualitative}
	\end{figure*}
	
	\begin{figure}[t]
		\centering
		\includegraphics[width=.9\linewidth]{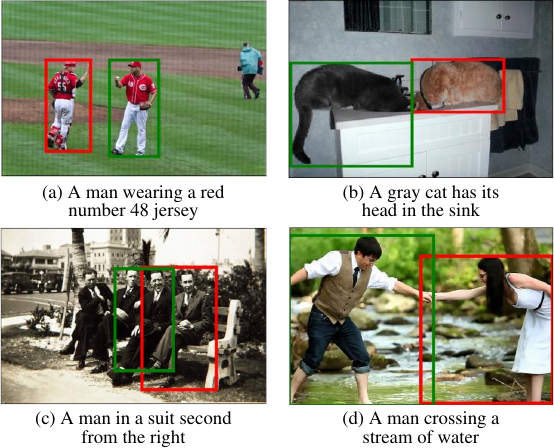}
		\caption{Some representative failure cases of our method. The red bounding-box represents the prediction of our method, and the green bounding-box is the corresponding ground-truth.}
		\label{fig:badcase}
	\end{figure}
	It is worth noting that all methods suffer forgetting when training is completed after task 4, especially on CRefCOCO and CRefCOCO+. We argue the reason is that the number of samples in task 4 is much more than in other tasks. The learning process breaks the balance of stability and plasticity.

	\begin{table}[t]
		\renewcommand\arraystretch{1.3}
		\renewcommand\tabcolsep{4.0pt}
		\caption{The average sub-module weight calculated by attention of DMM under 5-task setting. The sub-modules include subject, relation, location.}
		\label{tab:sub-module}
		\centering
		\begin{tabular}{cccc}
			\toprule
			Sub-Module & CRefCOCO & CRefCOCO+ & CRefCOCOg \\ \hline
			Subject & 0.51 & 0.66 & 0.57 \\
			Relation & 0.35 & 0.22 & 0.23 \\
			Location & 0.14 & 0.12 & 0.19 \\
			\bottomrule
		\end{tabular}
	\end{table}

	\subsection{Ablation Study}
	To deeply analyze our proposed DMM method, we study its different ablation variants on the re-split datasets. MAttNet is used as the backbone, and we conduct the experiment under the 5-task setting.
	
	\subsubsection{Effect of Different Variants}
	
	We first study the effectiveness of different variants of the Implicit-Memory and Explicit-Memory, including: i) Na\"ive Implicit-Memory (N-IM), ii) Weighted Implicit-Memory (W-IM), iii) Na\"ive Explicit-Memory (N-EM) and iv) Modular Explicit-Memory (M-EM).
	\eat{
		\begin{itemize}
			\item [(\romannumeral1)] Na\"ive Implicit-Memory (N-IM) constrains the parameters updating to alleviate catastrophic forgetting.
			\item [(\romannumeral2)] Weighted Implicit-Memory (W-IM) computes the module weighted to determine which module and parameters within the module to retain for a previous task.
			\item [(\romannumeral3)] Na\"ive Explicit-Memory (N-EM) denotes selecting samples only by ranking their loss.
			\item [(\romannumeral4)] Modular Explicit-Memory (M-EM) selects buffers by the multiplication of loss and the subject-weight calculated by the self-attention.
	\end{itemize}}
	The ablative results are described in Table \ref{tab:ablationtab}. 
	From the results, we can observe that each of the four variants brings consistent improvement on different benchmarks.
	In particular, both N-IM and N-EM outperform Finetuning to a large margin, which shows the effectiveness of our designed Implicit-Memory and Explicit-Memory for alleviating the stability-plasticity dilemma.
	We conduct the combination of N-IM and N-EM on our re-split datasets. The result indicates that Implicit-Memory and Explicit-Memory can compensate for each other to achieve a better result.
	When the N-IM and N-EM are imporved to W-IM and M-EM, respectively, the results are further enhanced. These enhancements validate the effectiveness of the W-IM and M-EM.
	In addition, the combination of W-IM and M-EM (i.e., DMM) achieves the best performance, 
	demonstrating the necessity of (1) considering the contribution of different sub-modules (for Implicit-Memory) and (2) using subject-module importance information to guide the memory updating  (for Explicit-Memory).

	\subsubsection{Effect of Different Sample-choosing Strategies}
	In our explicit-memory module, we select representative samples of a task by choosing those \textit{easy} samples leading to a small loss. 
	In this part, we study the impact of sample hardness on rehearsal performance. 
	Three strategies are compared, including
	(1) High-strategy (\textit{high}) chooses samples with the highest loss for explicit-memory;
	(2) Low-strategy (\textit{low}) chooses samples with the lowest loss;
	(3) Random-strategy (\textit{random}) performs sample selection randomly. 
	Table \ref{tab:em loss} shows the evaluation results on different datasets.
	As seen, it is the {Low}-strategy that achieves the best results,
	which confirms that the hardness of the selected sampling has an impact on the final performance and easy samples are more effective for the explicit-memory module than the hard ones.

	\subsubsection{Contribution of Different Memory size}
	\label{ablation-memorysize}
	As discussed in Section \ref{EM}, EM explicitly stores some samples which are representative in the buffer pool with memory size $K$.
	In this ablation study, we explore the sensitivity of our model to various $K$, including 80, 120 and 160.
	The average accuracy results are shown in Fig \ref{fig:memorysize}. 
	As shown in the figure, the results are roughly consistent on all three datasets, showing that our model is not sensitive to the memory size. We chose 120 by taking into account the performance as well as training time.
	
	\subsubsection{Contribution of Different Sub-modules}
	\label{ablation-submodule}
	As discussed in Section M-EM, M-EM takes  into  account  not  only  the loss  but  also how importance of the subject intrinsic information in each sample.
	In this ablation study, we evaluate the module weight of the \textit{subject}, \textit{relation} and \textit{location} of each sample.
	The average results of each dataset are shown in Table \ref{tab:sub-module}. 
	It is evident that the subject information is more important than the other counterparts, verifying our previous ideas.
	
	\subsection{Qualitative Results}
	The conclusion drawn in the quantitative analyses is confirmed by the qualitative evaluation reported in Fig. \ref{fig:qualitative}.
	The top row shows the training (image, expression) pair and the ground-truth bounding box from task 1.
	The bottom four rows represent results produced by different methods. 
	Each column denotes the comprehension result after training on each task. 
	We can make the following observations.
	Finetuning can correctly locate the object after task 1. 
	However, after task 3 is learned, it gradually forgets what a human is, indicating that it suffers from catastrophic forgetting.
	The weight implicit-memory method grounds the wrong object until the fourth task, showing that regularization contributes to preventing catastrophic forgetting.
	Furthermore, compared to weight implicit-memory, although modular explicit-memory does not always ground the correct object, it retains knowledge about task 1.
	Finally, DMM can ground correctly after learning of each task, further demonstrating the advantageous performance of our method.
	
	The Fig.\ref{fig:badcase} illustrates some failure cases. As shown in the Fig.\ref{fig:badcase}(a), we succeeded in locating the person however we do not get the number “48”.
	Other examples are shown in Fig.\ref{fig:badcase}(b) and Fig.\ref{fig:badcase}(c), the model loses the ability to capture the appearance and location information.
	In addition, after learning on multiple tasks, our model may lose some of its ability to acquire global information to discern the gender of a person, such as the case in Fig.\ref{fig:badcase}(d).
	We leave how to solve these failure cases as interesting future works.

	\section{Conclusions}
	In this work, we propose to study the continual referring comprehension problem. In this setting, each REC task aims to localize one object category and such REC tasks are presented to the model sequentially. To address the catastrophic forgetting problem in this continual problem, we proposed a novel and effective Dual Modular Memorization model. The model consists of two memory components. One component, termed Implicit-Memory module, learns to retain structural parameters of previous tasks. The other component, termed Explicit-Memory module, avoids forgetting previous tasks by retaining some representative samples of these tasks into a buffer, which will be replayed when learning new tasks. Experiments conducted on three datasets re-splited based on three benchmark REC datasets demonstrate the superiority of our model over a number of continual learning baselines. In this work, we assume there exist clear task boundaries between the tasks. In the future, we plan to go beyond this assumption and study continual REC in a more practical setting.
	
	\section*{Acknowledgment}
	This study was supported by grants from Chinese National Science \& Technology Pillar Program (No. 2022YFC2009900/2022YFC2009903), the National Natural Science Foundation of China (Grant No. 62122018, No. 62020106008, No. 61772116, No. 61872064).
	
	\eat{\subsection{Subsection Heading Here}
		Subsection text here.
		
		\subsubsection{Subsubsection Heading Here}
		Subsubsection text here.
		
		\section{Conclusion}
		The conclusion goes here.

		\appendices
		\section{Proof of the First Zonklar Equation}
		Appendix one text goes here.
		
		\section{}
		Appendix two text goes here.
		
		\section*{Acknowledgment}

		The authors would like to thank...}

	\ifCLASSOPTIONcaptionsoff
	\newpage
	\fi
	
	\bibliographystyle{IEEEtran}
	\bibliography{ref}
	
	\begin{IEEEbiography}[{\includegraphics[width=1in,height=1.25in,clip,keepaspectratio]{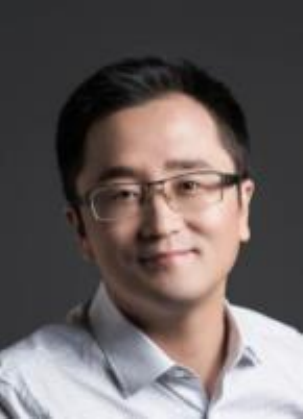}}]{Heng Tao Shen} is the Dean of School of Computer Science and Engineering, the Executive Dean of AI Research Institute at University of Electronic Science and Technology of China (UESTC). He obtained his BSc with 1st class Honours and PhD from Department of Computer Science, National University of Singapore in 2000 and 2004 respectively. His research interests mainly include Multimedia Search, Computer Vision, Artificial Intelligence, and Big Data Management. He is/was an Associate Editor of ACM Transactions of Data Science, IEEE Transactions on Image Processing, IEEE Transactions on Multimedia, IEEE Transactions on Knowledge and Data Engineering, and Pattern Recognition. He is a Member of Academia Europaea, Fellow of ACM, IEEE and OSA.  
	\end{IEEEbiography}
	
	\begin{IEEEbiography}[{\includegraphics[width=1in,height=1.25in,clip,keepaspectratio]{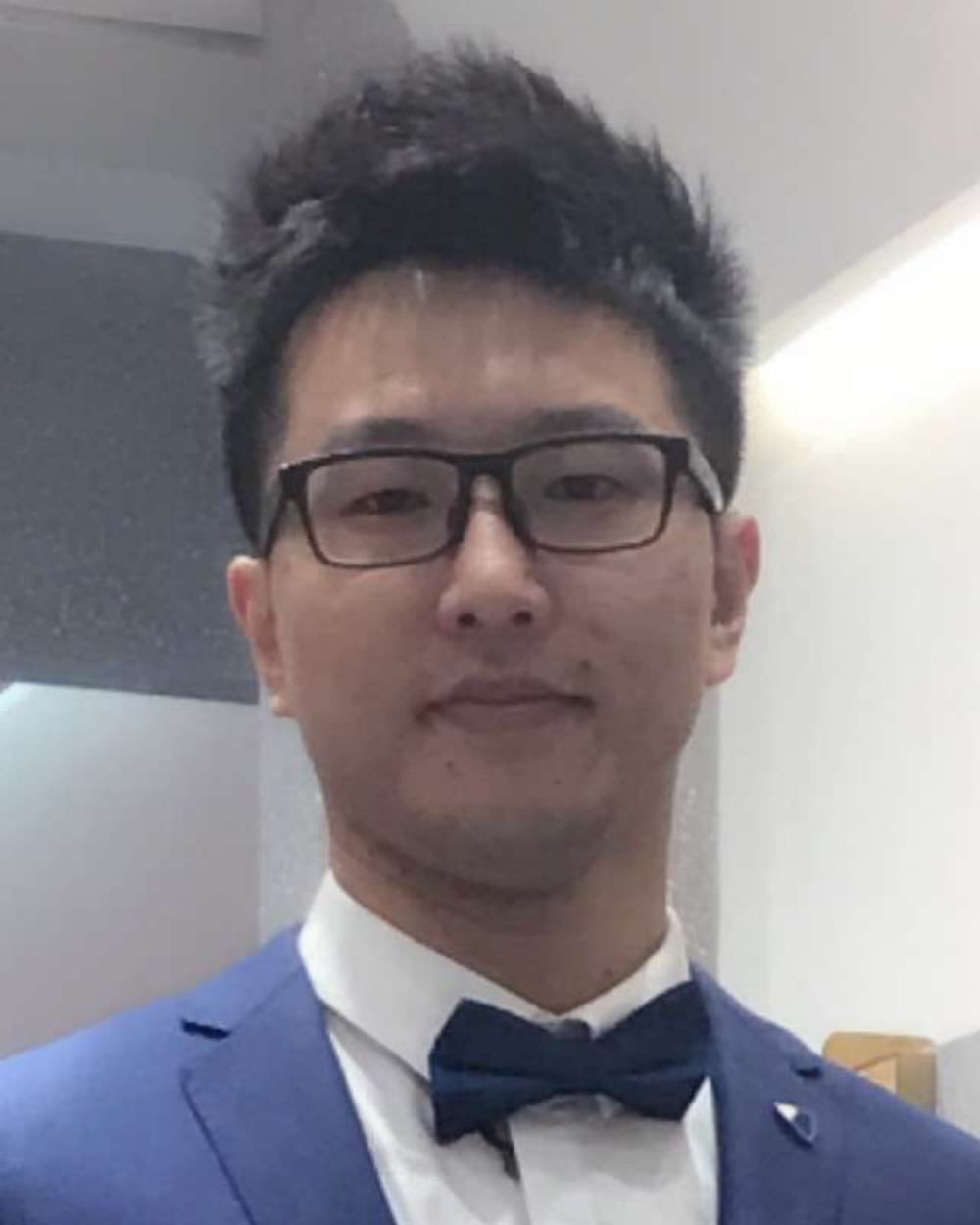}}]{Cheng Chen} is currently pursuing the Ph.D. degree with the School of Computer Science, University of Electronic Science and Technology of China, China. 
	
	His research interests include computer vision, continual learning, quantization.
	\end{IEEEbiography}
	
	\begin{IEEEbiography}[{\includegraphics[width=1in,height=1.25in,clip,keepaspectratio]{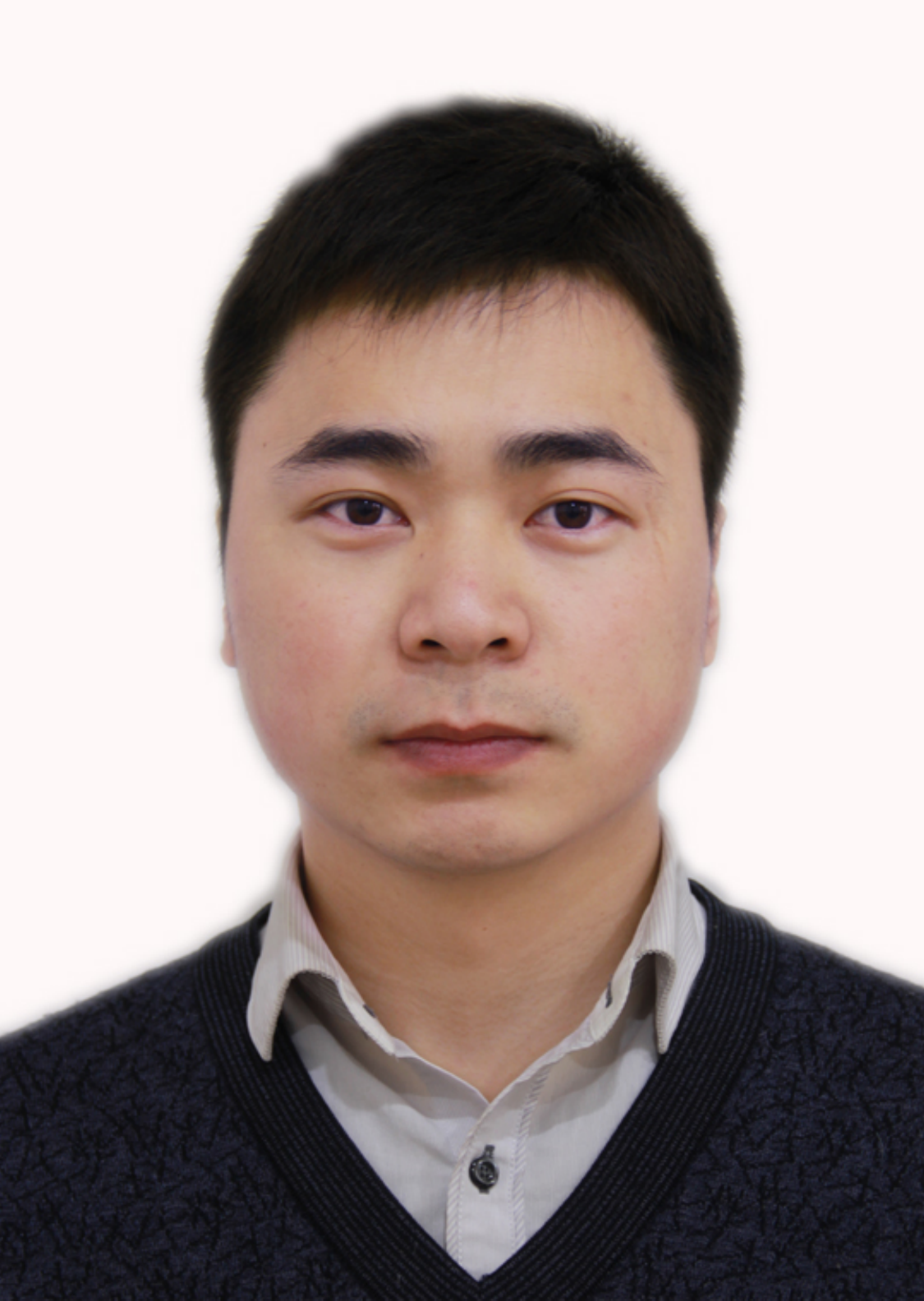}}]{Jingkuan Song} (Senior Member, IEEE) is currently a Professor with the University of Electronic Science and Technology of China (UESTC), Chengdu, China. His research interests include large-scale multimedia retrieval, image/video segmentation and image/video understanding using hashing, graph learning, and deep learning techniques. Dr. Song has been an AC/SPC/PC Member of IEEE Conference on Computer Vision and Pattern Recognition for the term 2018–2021, and so on. He was the winner of the Best Paper Award in International Conference on Pattern Recognition, Mexico, in 2016, the Best Student Paper Award in Australian Database Conference, Australia, in 2017, and the Best Paper Honorable Mention Award, Japan, in 2017.
	\end{IEEEbiography}
	
	\begin{IEEEbiography}[{\includegraphics[width=1in,height=1.25in,clip,keepaspectratio]{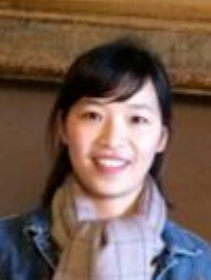}}]{Lianli Gao}(Member, IEEE) received the Ph.D. degree in information technology from The University of Queensland (UQ), Brisbane, QLD, Australia, in 2015. She is currently a Professor with the School of Computer Science and Engineering, University of Electronic Science and Technology of China (UESTC), Chengdu, China. She is focusing on integrating natural language for visual content understanding. Dr. Gao was the winner of the IEEE Trans. on Multimedia 2020 Prize Paper Award, the Best Student Paper Award in the Australian Database Conference, Australia, in 2017, the IEEE TCMC Rising Star Award in 2020, and the ALIBABA Academic Young Fellow.
	\end{IEEEbiography}
	
	\begin{IEEEbiography}[{\includegraphics[width=1in,height=1.25in,clip,keepaspectratio]{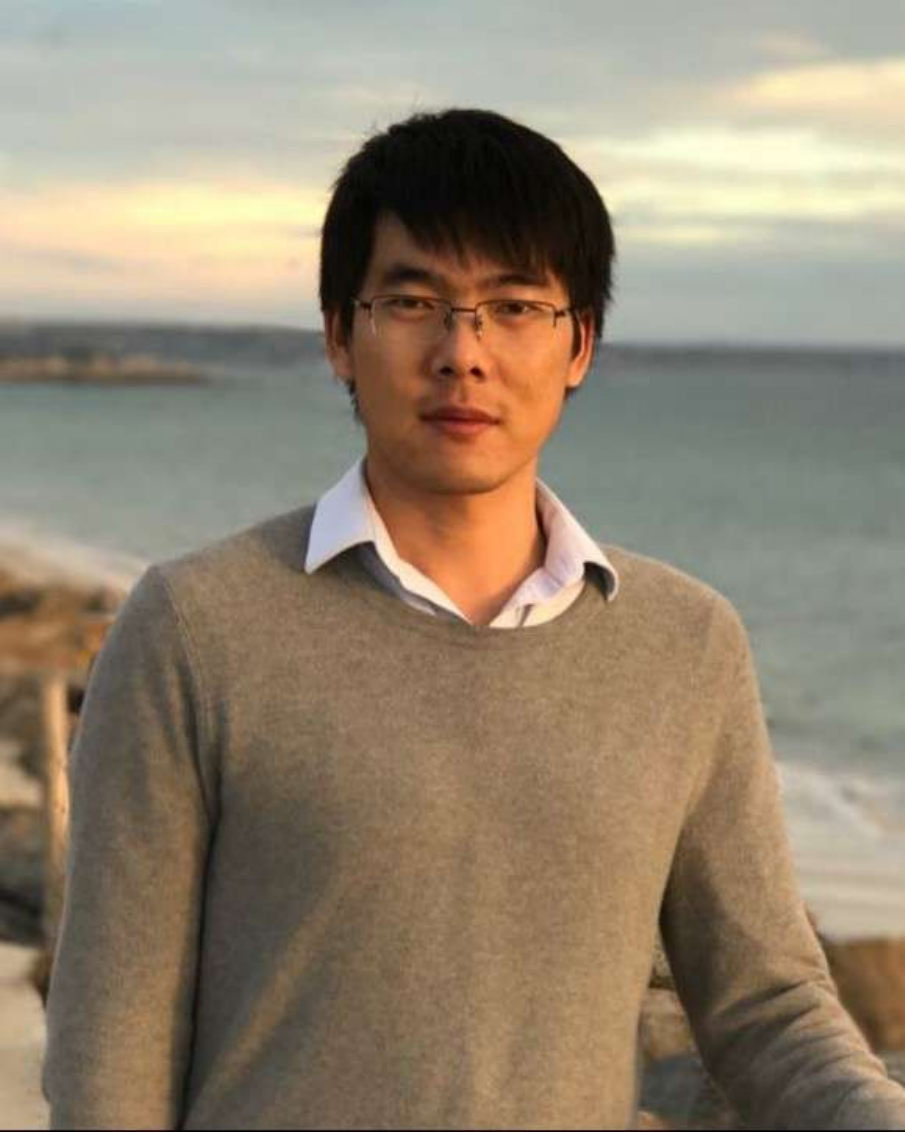}}]{Peng Wang} received his Ph.D. degree from School of Information Technology and Electrical Engineering, University of Queensland in 2017. He is a lecturer with School of Computing and Information Technology, University of Wollongong (UOW). Before joining UOW, he was a research fellow with Australian Institute for Machine Learning. His research interest lies in computer vision and deep learning. He has been actively publishing his research work on top-tier international journals and conferences, such as IEEE TPAMI, IJCV, CVPR, ECCV, etc. He will serve as workshop chair for ACM Multimedia 2021 and tutorial chair for ACCV 2022. He is regular reviewer of IEEE TPAMI, CVPR, ICCV, ECCV, etc. He was recognized as outstanding reviewer by ECCV 2020.
	\end{IEEEbiography}
	
	\begin{IEEEbiography}[{\includegraphics[width=1in,height=1.25in,clip,keepaspectratio]{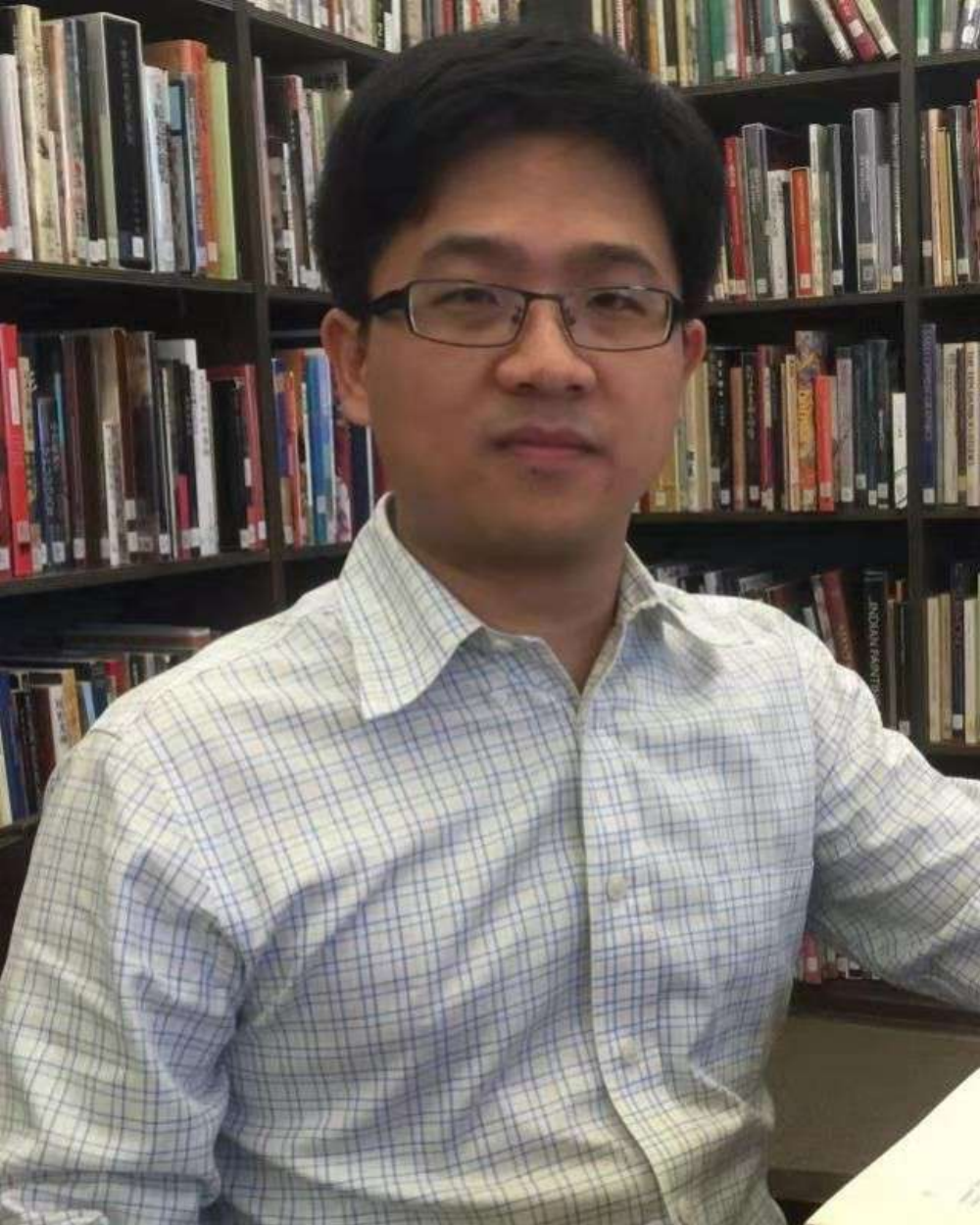}}]{Meng Wang} (Fellow, IEEE) received the B.E. and Ph.D. degrees in the special class for the gifted young from the Department of Electronic Engineering and Information Science, University of Science and Technology of China, Hefei, China, in 2003 and 2008, respectively. He is currently a Professor with the Hefei University of Technology, Hefei. He has authored over 200 book chapters, journal and conference papers in his research areas. His current research interests include multimedia content analysis, computer vision, and pattern recognition. Prof. Wang was a recipient of the ACM SIGMM Rising Star Award 2014. He is an Associate Editor of the IEEE TRANSACTIONS ON KNOWLEDGE AND DATA ENGINEERING, IEEE TRANSACTIONS ON CIRCUITS AND SYSTEMS FOR VIDEO TECHNOLOGY, and IEEE TRANSACTIONS ON NEURAL NETWORKS AND LEARNING SYSTEMS.
	\end{IEEEbiography}
	
\end{document}